\documentclass[
twocolumn
]{ceurart}

\usepackage{fancyvrb}
\usepackage{balance}
\usepackage{listings}
\usepackage{graphicx}
\usepackage[export]{adjustbox}
\usepackage[relative,overlay]{textpos}
\usepackage{tikz}
\newcommand*\circled[1]{\tikz[baseline=(char.base)]{
    \node[shape=circle,fill=black, text=white,draw,inner sep=2pt] (char) {#1};}}

\begin{document}

\copyrightyear{2023}
\copyrightclause{Copyright for this paper by its authors.
  Use permitted under Creative Commons License Attribution 4.0
  International (CC BY 4.0).}

\conference{Proceedings of the Sixth Workshop on Automated Semantic Analysis of Information in Legal Text (ASAIL 2023), June 23, 2023, Braga, Portugal.}

\title{Can GPT-4 Support Analysis of Textual Data in Tasks Requiring Highly Specialized Domain Expertise?}


\author[1]{Jaromir Savelka}[%
orcid=0000-0002-3674-5456,
email=jsavelka@cs.cmu.edu,
url=https://www.cs.cmu.edu/~jsavelka/,
]
\cormark[1]
\address[1]{Computer Science Department, Carnegie Mellon University, Pittsburgh, PA, USA}

\author[2]{Kevin D. Ashley}[%
email=ashley@pitt.edu,
]
\author[2]{Morgan A. Gray}[%
orcid=0000-0002-3800-2103,
email=mag454@pitt.edu,
]
\address[2]{Intelligent Systems Program, University of Pittsburgh, PA, USA}

\author[3]{Hannes Westermann}[
orcid=0000-0002-4527-7316,
email=hannes.westermann@umontreal.ca
]
\address[3]{Cyberjustice Laboratory, Faculté de droit, Université de Montréal, Montréal, Canada}

\author[2]{Huihui Xu}[%
email=huihui.xu@pitt.edu,
]

\cortext[1]{Corresponding author.}

\begin{abstract}
  We evaluated the capability of generative pre-trained transformers~(GPT-4) in analysis of textual data in tasks that require highly specialized domain expertise. Specifically, we focused on the task of analyzing court opinions to interpret legal concepts. We found that GPT-4, prompted with annotation guidelines, performs on par with well-trained law student annotators. We observed that, with a relatively minor decrease in performance, GPT-4 can perform batch predictions leading to significant cost reductions. However, employing chain-of-thought prompting did not lead to noticeably improved performance on this task. Further, we demonstrated how to analyze GPT-4's predictions to identify and mitigate deficiencies in annotation guidelines, and subsequently improve the performance of the model. Finally, we observed that the model is quite brittle, as small formatting related changes in the prompt had a high impact on the predictions. These findings can be leveraged by researchers and practitioners who engage in semantic/pragmatic annotations of texts in the context of the tasks requiring highly specialized domain expertise.
\end{abstract}


\maketitle

\section{Introduction}
This paper assesses the capability of generative pre-trained transformers (GPT), specifically OpenAI's GPT-4, to automatically perform semantic analysis of sentences extracted from court opinions \cite{savelka2017sentence} to support interpretation of legal concepts as used in statutory law. The multi-label sentence classification task requires highly specialized legal domain expertise. We use selected parts of an existing manually labeled data set\footnote{Statutory Interpretation Data Set. Available at: \url{https://github.com/jsavelka/statutory_interpretation} [Accessed 2023-05-01]} to assess the effectiveness of GPT-4, comparing it to the performance of human annotators. Further, we explore the implications of processing the data in batches as a cost effective alternative to analyzing one data point at a time. We also report the results of our prompt engineering efforts aimed at improving the effectiveness of the system on the task. These include general techniques, such as chain of thought prompting (CoT) \cite{wei2022chain}, as well as task specific tweaking of annotation guidelines. Finally, we assess GPT-4's predictions in terms of their robustness.

Early systematic efforts of applying empirical methods of computational linguistics to semantic, discourse-related, and/or pragmatic aspects of textual data date back to the mid 1990s \cite{artstein2008inter}. Such efforts require annotated resources which have traditionally relied on subjective human judgement. In the legal domain, the approach has been embraced in many practical workflows in eDiscovery or contract review as well as in the research field of empirical legal analysis.  Machine learning (ML) methods enabled  approaches where humans needed to annotate only a part of the  corpus.  The remainder is analyzed automatically via a ML system trained on the manually annotated portion of the data. 

Recently, a new paradigm has emerged where a large language model (LLM) is employed in  zero/few-shot settings, using carefully crafted natural language prompts (akin to human readable instructions) \cite{liu2023pre}. This paradigm could be valuable because it may enable generation of high quality annotations with less demand for human annotators---an expensive resource, especially in tasks that require highly specialized domain expertise. Such expertise is often required in analysis of legal documents such as court opinions or statutory provisions. The cost of human labor required to annotate large legal data sets has been an important bottleneck in carrying out certain types of research in the field of AI \& Law.

To investigate the capability of GPT-4 to analyze court opinions in the context of the task focused on interpretation of legal concepts from statutory law, we analyzed the following research questions:

\begin{enumerate}
    \item[(RQ1)] How successfully can GPT-4 perform the task as compared to human annotators?
    \item[(RQ2)] Can GPT-4 perform the task as batch prediction, i.e., analyzing multiple data points at the same time?
    \item[(RQ3)] Does the accuracy of GPT-4's predictions improve when the model is forced to provide explanations (akin to CoT)?
    \item[(RQ4)] What are the effects of modifying the annotation guidelines based on the identified shortcomings?
    \item[(RQ5)] How robust (i.e., stable) are the predictions of GPT-4 against changes of the prompt that are not related to the task definition?
\end{enumerate}

By carrying out this work, we provide the following contributions to the AI \& Law research community. As far as we know, this is the first study that, in the context of a task requiring highly specialized legal expertise:

\begin{enumerate}
    \item[(C1)] Benchmarks the performance of human annotators to the performance of GPT-4 prompted with an (almost) exact copy of annotation guidelines.
    \item[(C2)] Compares the performance of GPT-4 on batch prediction to the performance of analyzing a single data point at a time.
    \item[(C3)] Reports and discusses results of diverse prompt engineering efforts aimed at improving task specific performance of GPT-4.
    \item[(C4)] Analyzes the robustness of GPT-4's predictions.
\end{enumerate}

\section{Related Work}
LLMs have shown promising results in various text analysis tasks. Wang et al. \cite{wang_want_2021} and Ding et al. \cite{ding_is_2022} explored the use of GPT-3 for data labeling in tasks such as text entailment, sentiment analysis, topic classification, summarization, question generation, or named entity recognition. Multiple studies demonstrated that ChatGPT outperforms crowd-workers in text annotation tasks \cite{gilardi_chatgpt_2023,tornberg_chatgpt-4_2023}. At the same time, researchers caution about issues with reliability of ChatGPT in such tasks \cite{reiss_testing_2023}. There are several studies employing various GPT models to analyze texts within tasks that require specialized domain expertise. For example, Kuzman et al. examined ChatGPT on the task of automatic genre identification \cite{kuzman_chatgpt_2023}. Huang et al. investigated the strengths and limitations of ChatGPT in annotating implicit hate speech \cite{huang_is_2023}. Ziems et al. discussed the potential of LLMs to transform computational social science and the role they could play in social science analysis \cite{ziems_can_nodate}. Zhu et al. explored ChatGPT's capabilities in reproducing human-generated label annotations in social computing tasks \cite{zhu_can_2023}. Our study explores the efficacy of GPT-4 for analysis of texts of court opinions in the context of the task focused on interpretation of legal concepts from statutory law.

This work explores the use of GPT-4 to support semantic analysis of legal texts. There has been a growing interest in exploring capabilities of GPT models in such applications. Yu et al. applied GPT-3 to the COLIEE legal entailment task that is based on the Japanese Bar exam, substantially improving over the state-of-the-art result \cite{https://doi.org/10.48550/arxiv.2212.01326}. Similarly, Bommarito and Katz utilized GPT-3.5 for the Multistate Bar Examination \cite{bommarito2022gpt}. Later, Katz et al. applied GPT-4 to the entire Uniform Bar Examination (UBE) and observed the system passing the exam \cite{katz2023gpt}. Other use cases involve assessment of trademark distinctiveness \cite{goodhue2023classification}, legal reasoning \cite{blair2023can,nguyen2023well}, including statutory interpretation \cite{savelka2023concepts}, U.S. Supreme court judgment modeling \cite{hamilton2023blind}, providing legal information \cite{tan2023}, annotation of legal documents \cite{savelka2023unlocking}, and online dispute resolution \cite{westermann2023llmediator}.

A steady line of work in AI \& Law focuses on making the text analysis effort (i.e., annotation) more effective. Westermann et al. proposed and assessed a method for building strong, explainable classifiers in the form of Boolean search rules \cite{westermann2019}, as well as a method based on sentence semantic similarity \cite{westermann2020}. Savelka and Ashley evaluated the effectiveness of an approach where a user labels the documents by confirming (or correcting) the prediction of a ML algorithm \cite{savelka2015}. The application of active learning has been explored in the context of classification of statutory provisions \cite{waltl2017} and eDiscovery \cite{cormack2016,cormack2015}. Hogan et al. proposed and evaluated a human-aided computer cognition framework for eDiscovery \cite{hogan2009}. In this study, we evaluate the zero-shot capabilities of GPT-4 to support the analysis.

\section{Data}
\label{sec:data}
To investigate the research questions listed above, we use a subset of the data set released in \cite{savelka2021discovering} focused on interpretation of legal concepts from statutory provisions. Statutory and regulatory provisions are difficult to understand because the rules they express must account for diverse situations, even those not yet encountered. When the application of a general rule is not straightforward a lawyer must present arguments as to why a provision should be applied in a particular way. In doing so the lawyer must often defend a specific account of the meaning of one or more terms (i.e. ``phrase of interest''). A thorough analysis of the past treatment of the phrase of interest is foundational to formation of an adequate argument. The treatment consists of past mentions and uses of the phrase in sentences from documents such as court decisions, legislative histories, or journal articles.

\begin{figure*}[t]
  \centering
  \label{fig:mockup_interface}
  \includegraphics[width=\linewidth,frame]{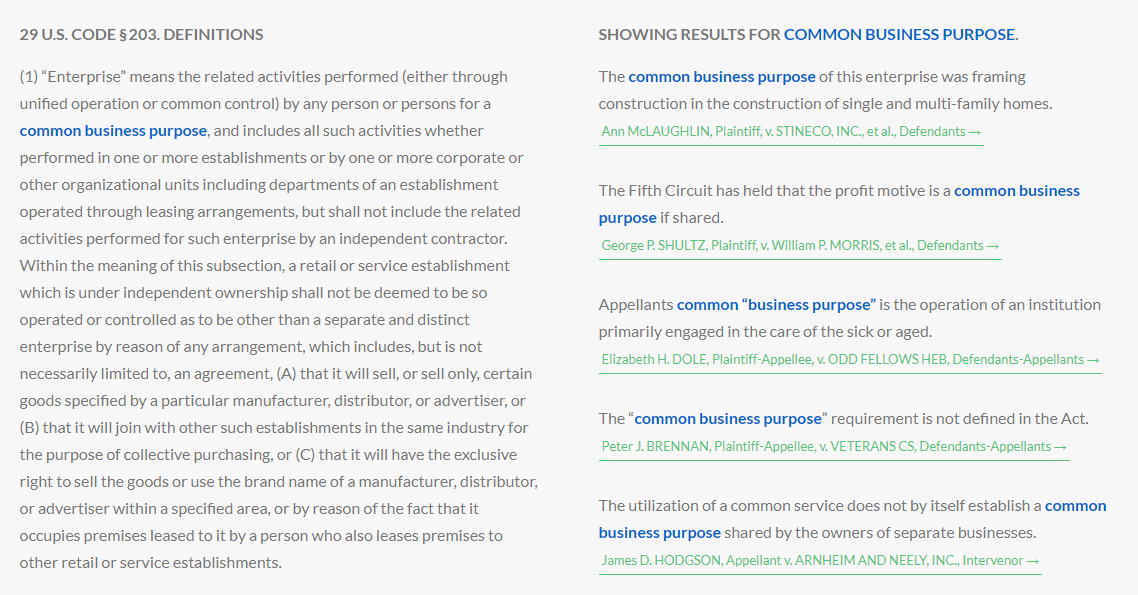}
  \begin{textblock*}{3.4cm}(2cm,-7.5cm)\circled{1}\end{textblock*}
  \begin{textblock*}{3.4cm}(1.3cm,-6.25cm)\circled{2}\end{textblock*}
  \begin{textblock*}{3.4cm}(5.75cm,-6.8cm)\circled{3}\end{textblock*}
  \begin{textblock*}{3.4cm}(11.4cm,-6.25cm)\circled{4}\end{textblock*}
  \caption{A mock-up interface with an example statutory provision on the left (1). The user indicated that they are interested in the meaning of the ``common business purpose`` phrase as used in the provision (2). The system responds with a list of sentences that are deemed useful for explaining the meaning of the phrase (e.g., 3). The user may follow the link to the full-text of an opinion to view the sentence in its original context (4).}
\end{figure*}

The ability to sift through large amounts of legal documents and distill the content, that could be subsequently used in argumentation about the meaning of a phrase, is an important part of any lawyer's skill set. To understand the value of a sentence that uses the phrase of interest one may need to answer questions such as:

\begin{itemize}
\item Does a sentence provide additional information to what is already known from the statutory provision?
\item Does the sentence content provide solid grounds for understanding some useful facets of the meaning of the phrase of interest?
\item Is the meaning of the phrase used in the sentence the same as the meaning of the phrase of interest?
\end{itemize}

Given a text of a single statutory provision (i.e., the source provision) and the phrase of interest (i.e., one or more words in whose meaning we are interested), the task is to evaluate sentences' as to their \textit{explanatory value} \cite{savelka2022role}. The sentences come from case decisions responsive to a query in the form of the phrase of interest (e.g. ``common business purpose''). A sentence should be labeled with one of the following categories \cite{vsavelka2016extracting}:

\begin{itemize}
\item \textbf{High value}: This label is reserved for sentences that explicitly elaborate on the meaning of the phrase of interest.
\item \textbf{Certain value}: The system should select this label if the sentence does not explicitly elaborate on the meaning of the phrase of interest, yet the sentence still provides grounds to draw some (even modest or quite vague) conclusions about the meaning of the phrase of interest.
\item \textbf{Potential value}: This label is appropriate if the sentence does not appear to be useful for elaboration on the meaning of the phrase of interest but the sentence provides some additional information (even quite marginal) over what is known from the source provision.
\item \textbf{No value}: This label should be selected if the sentence does not provide any additional useful information over what is already known from the source provision.
\end{itemize}

\noindent This type of text analysis may enable training of ML models supporting, e.g., a legal information retrieval system focused on legal concepts interpretation such as the one shown in Figure \ref{fig:mockup_interface} \cite{savelka2019improving,savelka2020learning,vsavelka2021legal}.

The original data set was annotated by domain experts---11 law students and 2 legal scholars with law degrees. The law students performed the first pass of the annotations and the scholars were responsible for the second pass resulting in the consensus labels. The agreement between the students' annotations and the consensus labels, measured in terms of Krippendorff's $\alpha$ \cite{Krippendorff2011}, was $0.1<\alpha<0.6$ (see Figure \ref{fig:alpha}) while the inter-annotator agreement between the two scholars was $\alpha=0.79$ \cite{savelka2020discovering}. Hence, clearly this is a very demanding text analysis task requiring highly specialized domain expertise.

\begin{table}[t]
  \caption{Data set descriptive statistics showing the distribution of sentence labels per phrase of interest (the first column). NV -- No value, PV -- Potential value, CV -- Certain value, HV -- High value.}
  \label{tab:data_set}
  \setlength{\tabcolsep}{3pt}
  \begin{tabular}{lrrrrr}
  \toprule
    Phrase of interest            & NV & PV & CV & HV & Total \\
  \midrule
    Accommodation trade           & 4  & 48 & 10 & 7  & 69 \\
    Cybercrime sentence           & 4  & 54 & 11 & 2  & 71 \\
    Digital musical recording     & 6  & 13 & 11 & 13 & 43 \\
    Semiconductor chip product    & 2  & 9  & 12 & 2  & 25 \\
    Unduly disrupt the operations & 7  & 36 & 2  & 3  & 48 \\
  \midrule
    Total                         & 23 &160 & 46 & 27 &256 \\
  \bottomrule
\end{tabular}
\end{table}

The original data set consists of 42 queries (i.e., phrases of interest) associated with 26,959 labeled sentences from 20 different areas of legal regulation (e.g., intellectual property, criminal law). Considering the non-negligible cost of large numbers of requests to the GPT-4 API, we decided to work with a small subset of the original data set. We selected 5 phrases of interest associated with 256 sentences. While limited, the sample of this size is sufficient to support the experiments in this work. The distribution of labels within the data set is reported in Table \ref{tab:data_set}.

\section{Model}
In our experiments, we use the GPT-4 model. As of the writing of this paper, GPT-4 is by far the most advanced model released by OpenAI. The model is focused on dialog between a user and a system (i.e., an assistant).  The original GPT model \cite{radford2018improving} is a 12-layer decoder-only transformer \cite{vaswani2017attention} with masked self-attention heads. Its core capability is fine-tuning on a downstream task. The GPT-2 model~\cite{radford2019language} largely follows the details of the original GPT with a few modifications, such as layer normalization moved to the input of each sub-block, additional layer-normalization after the first self-attention block, and a modified initialization. Compared to the original model it displays remarkable multi-task learning capabilities~\cite{radford2019language}. The third generation of GPT models~\cite{brown2020language} uses almost the same architecture as GPT-2. The only difference is that it alternates dense and locally banded sparse attention patterns in the layers of the transformer. The main focus of \cite{brown2020language} was to study the dependence of performance and model size where eight differently sized models were trained (from 125 million to 175 billion parameters). The largest of these models is commonly referred to as GPT-3. The interesting property of these models is that they appear to be very strong zero- and few-shot learners. This ability appears to improve with the increasing size of the model~\cite{brown2020language}. The technical details of the recently released GPT-4 model have not been disclosed due to concerns about potential misuses of the technology as well as a highly competitive market for generative AI \cite{openai2023gpt4}.

We set the \verb|temperature| of the model to 0.0, which corresponds to no randomness. The higher the \verb|temperature| the more creative the output but it can also be less factual. As the temperature approaches 0.0, the model becomes deterministic and can be repetitive. We set \verb|max_tokens| to various values depending on the expected size of the output (a token roughly corresponds to a word) as this parameter controls the maximum length of the completion (i.e., the output). For a single data point classification task where we only expect a single label as a completion the setting of 50 is sufficient. For a batch classification with a CoT prompt, a much larger size of output is expected (1,500 tokens). Note that GPT-4 has an overall token length limit of 8,192 tokens, comprising both the prompt and the completion.\footnote{There is also a variant of the model that supports up to 32,768 tokens.} We set \verb|top_p| to 1, as is recommended  when \verb|temperature| is set to 0.0. This parameter is related to \verb|temperature| and also influences creativeness of the output. We set \verb|frequency_penalty| to 0, which allows repetition by ensuring no penalty is applied to repetitions. Finally, we set \verb|presence_penalty| to 0, ensuring no penalty is applied to tokens appearing multiple times in the output.

\section{Experimental Design}

\subsection{GPT-4 Text Analysis (RQ1)}
\label{sec:rg1}
The first experiment was focused on answering RQ1, i.e., how successfully GPT-4 can perform the annotation task, as compared to human annotators. To that end we used the annotation guidelines\footnote{Annotation Guidelines for Evaluating Sentences for Argumentation about the Meaning of Statutory and Regulatory Terms. Available at: \url{https://github.com/jsavelka/statutory_interpretation/blob/master/annotation_guidelines_v2.pdf}  [Accessed 2023-04-30]} originally designed for the human annotators and turned them into a system prompt for GPT-4. The system prompt is typically used to steer the system (i.e., the GPT-4 model) towards performing the desired task. We introduced only minimal changes to the annotation guidelines to ensure close mapping between the original task performed by human annotators and the task performed by GPT-4 automatically. We left out pieces of the annotation guidelines related to the specifics of the annotation environment used by humans, as these would have made no sense in the GPT-4's prompt, e.g.:

\begin{quote}
At the top of each sheet there is a cell with a light
yellow background that contains a text of a single statutory provision [...]
\end{quote}

\noindent Furthermore, we replaced  references to ``students'' with a reference  to a ``system''. The guidelines contained a visual diagram, encoding the workflow of annotation rules which we translated into a list of questions. 
Finally, we omitted several examples in order to fit the annotation guidelines within the prompt and leave sufficient space for the output. The overall structure of the system prompt (i.e., the annotation guidelines) is shown in Figure \ref{fig:system-prompt}. Note that this  sizeable piece of text is much longer than what is typically used as a system prompt with GPT-4.

\begin{figure}
\footnotesize
\begin{Verbatim}[frame=single,commandchars=\\\{\}]
You are a specialized system focused on semantic 
annotation of court opinions.

BACKGROUND
Statutory and regulatory provisions are diff...
\textcolor{gray}{[3,300 characters ...]}

ANNOTATION TASK
The system is provided with a text of a single
\textcolor{gray}{[1,508 characters ...]}

RULES FOR SENTENCE EVALUATION
The system should evaluate the sentence using
\textcolor{gray}{[5,648 characters ...]}
\end{Verbatim}
\begin{textblock*}{3.4cm}(-0.65cm,-4.1cm)
\circled{1}
\end{textblock*}
\begin{textblock*}{3.4cm}(-1.35cm,-3.85cm)
\circled{2}
\end{textblock*}
\begin{textblock*}{3.4cm}(-0.65cm,-2.58cm)
\circled{3}
\end{textblock*}
\begin{textblock*}{3.4cm}(1.4cm,-1.3cm)
\circled{4}
\end{textblock*}
\caption{The system Prompt is populated with annotation guidelines as shown above. The typical preamble (1) is followed by the Background section (2) describing the context of the text analysis task to be performed. The Annotation Task section (3) provides more specific information about the mechanics of the task. Finally, the Rules for Sentence Evaluation section (4) contains the fine-grained instructions on how to categorize retrieved sentences. The grey tokens inform about the size of the parts of the prompt not shown in the figure. The prompt is a sizeable text spanning multiple pages.}
\label{fig:system-prompt}
\end{figure}

Each data point was provided to the system as a message coming from a user. The message contained the phrase of interest, citation to the source provision, the text of the source provision, as well as a retrieved sentence that should have been labeled with one of the categories described in Section \ref{sec:data}. The exact layout and formatting of the message is provided in Figure \ref{fig:user_msg}. GPT-4 was expected to return a message (coming from an assistant) containing the predicted label. In this experiment we set the \verb|max_tokens| parameter to 50 as this was sufficient for this type of completion.

\begin{figure}
\footnotesize
\begin{Verbatim}[frame=single,commandchars=\\\{\}]
PHRASE OF INTEREST: \textcolor{blue}{\string{\string{phrase_of_interest\string}\string}}

SOURCE PROVISION:
\textcolor{blue}{\string{\string{source_provision_citation\string}\string}}
\textcolor{blue}{\string{\string{source_provision_text\string}\string}}

SENTENCE:
\textcolor{blue}{\string{\string{sentence\string}\string}}

EXPECTED OUTPUT FORMAT:
Label: <label>
\end{Verbatim}
\begin{textblock*}{3.4cm}(0.45cm,-1.0cm)
\circled{\#}
\end{textblock*}
\caption{User message template for a single sentence prediction. The tokens shown in blue and surrounded by double curly braces are replaced with the corresponding data elements. Hence, the message is typically a somewhat longer text. Note the Expected Output Format section (\#) instructing the model as to the expected format of the response.}
\label{fig:user_msg}
\end{figure}

We inserted each data point from the data set into the template from Figure \ref{fig:user_msg} and submitted it individually to OpenAI's GPT-4 API, together with the system prompt. Note that this approach, despite the limited size of the data set of 256 samples, incurred a non-negligible cost exceeding \$20. The cost was, of course, lower than the cost of equivalent human labor on the same task. We extracted the predicted labels from the GPT-4 responses and compared them to the gold labels (Section \ref{sec:results}).

\subsection{Batch Prediction (RQ2)}
The next experiment was focused on answering RQ2, that is, whether GPT-4 can perform the task as batch prediction. To this end we used the same system prompt as in the preceding experiment (Figure \ref{fig:system-prompt}). We modified the user message as shown in Figure \ref{fig:user_msg_batch}. Instead of a single data point (i.e., sentence), we inserted multiple sentences. Correspondingly, the expected output part of the message was changed to reflect that GPT-4 should have returned more than one prediction. We constructed the batches dynamically to fit as many sentences as possible using the \verb|tiktoken| Python library\footnote{tiktoken. Available at: \url{https://github.com/openai/tiktoken} [Accessed: 2023-04-30]} to determine the size of the prompt before sending it to the GPT-4 API. Hence, the size of each batch is determined by the length of the submitted sentences. Typically, several tens of sentences were submitted within a single batch. For this experiment, we increased the \verb|max_tokens| parameter to 1,000 to accommodate lengthier completions. Note that this approach was significantly cheaper than the one presented earlier.

\begin{figure}
\footnotesize
\begin{Verbatim}[frame=single,commandchars=\\\{\}]
\textcolor{gray}{[...]}
SENTENCES:
Sentence 1: \textcolor{blue}{\string{\string{sentence_1\string}\string}}
Sentence 2: \textcolor{blue}{\string{\string{sentence_2\string}\string}}
\textcolor{gray}{[...]}

EXPECTED OUTPUT FORMAT:
Sentence 1: <label>
Sentence 2: <label>
Sentence 3: <label>
\end{Verbatim}
\begin{textblock*}{3.4cm}(0.45cm,-1.65cm)
\circled{\#}
\end{textblock*}
\caption{Excerpt from the user message template for batch prediction. The top part of the template that is omitted from the figure is the same as that shown in Figure \ref{fig:user_msg}. The tokens shown in blue and surrounded by double curly braces are replaced with the corresponding data elements. Note the Expected Output Format section (\#) instructing GPT-4 how to output multiple labels related to the submitted sentences.}
\label{fig:user_msg_batch}
\end{figure}

\subsection{Explanations -- CoT (RQ3)}
To explore RQ3, i.e., the effects of requiring the model to explain its predictions, we first modified the user message submitted to GPT-4 as shown in Figure \ref{fig:user_msg_exp}. This experiment was similar to the first one. The only difference was that we asked the model to first spell out an explanation regarding the predicted label, and to provide the prediction after that. This was inspired by the work on chain of thought (CoT) prompting that has been shown to improve performance of the models on diverse tasks \cite{wei2022chain}, including those in the legal domain \cite{https://doi.org/10.48550/arxiv.2212.01326}. For this experiment, we set the \verb|max_tokens| parameter to 500 to accommodate the expected completions. Given the increased size of the completion, this approach was even costlier than the one presented as the first experiment.

\begin{figure}
\footnotesize
\begin{Verbatim}[frame=single,commandchars=\\\{\}]
\textcolor{gray}{[...]}
EXPECTED OUTPUT FORMAT:
Explanation: <reasoning why particular label  
              should be assigned>
Label: <label>
\end{Verbatim}
\begin{textblock*}{3.4cm}(0.45cm,-1.65cm)
\circled{\#}
\end{textblock*}
\caption{Excerpt from the user message template requiring explanation before prediction. The top part of the template that is omitted from the figure is the same as that shown in Figure \ref{fig:user_msg}. Note the Expected Output Format section (\#) instructing GPT-4 how to output the explanation before the prediction.}
\label{fig:user_msg_exp}
\end{figure} 

To further explore RQ3, we modified the user message as shown in Figure \ref{fig:user_msg_batch_exp}. Here, we tested the effects of requiring explanations in the batch predictions task. Since full-blown natural language explanations, as in the preceding experiment, would have drastically decreased the size of the batch that could have been submitted to the API, we opted for schematic explanations encoding the answers of the model to the individual questions from the annotation guidelines stemming from the visual workflow (see Section \ref{sec:rg1}). For this experiment, we increased the \verb|max_tokens| parameter to 1,500 tokens. Note that this experiment was slightly more costly than the original batch prediction experiment. This was because GPT-4's completions cost more than the tokens submitted to the API. However, the cost was still significantly reduced when compared to the two experiments where the data points are submitted one by one.

\begin{figure}
\footnotesize
\begin{Verbatim}[frame=single,commandchars=\\\{\}]
\textcolor{gray}{[...]}
EXPECTED OUTPUT FORMAT:
Sentence 1: <explanation> => <label>
Sentence 2: <explanation> => <label>
Sentence 3: <explanation> => <label>

EXAMPLE:
Sentence 1: Q1 No => no value
Sentence 2: Q1 Yes -> Q2 No -> Q4 Yes => high 
            value
Sentence 3: Q1 Yes -> Q2 No -> Q4 No -> Q5 No => 
            potential value
...
\end{Verbatim}
\begin{textblock*}{3.4cm}(0.45cm,-4.15cm)
\circled{1}
\end{textblock*}
\begin{textblock*}{3.4cm}(-1.7cm,-2.55cm)
\circled{2}
\end{textblock*}
\caption{Excerpt from the user message template requiring explanations before predictions (batch). The top part of the template that is omitted from the figure is the same as that shown in Figure \ref{fig:user_msg_batch}. Note the Expected Output Format section (1) instructing GPT-4 how to output the schematic explanations before the predictions as well as the Examples section (2).}
\label{fig:user_msg_batch_exp}
\end{figure}

\subsection{Prompt (Annotation Guidelines) Modification (RQ4)}
The next experiment was focused on answering RQ4, i.e., analyzing the effects of modifying the annotation guidelines. In a typical annotation workflow where human annotators are involved, the early stages are dedicated to the training of the human annotators as well as to the refinement of annotation guidelines. Note that GPT-4 provides means for similar types of interventions. The training of the human annotators is akin to augmenting GPT-4's prompt with labeled examples (i.e., few-shot settings) or fine-tuning the model. The refinement of annotation guidelines translates into modifications of the system prompt containing the guidelines. In this work, we focused on exploring the refinement of annotation guidelines (i.e., the prompt), leaving the exploration of few-shot learning and fine-tuning as open questions for future work.

Based on the results of the preceding four experiments, we identified a prominent weakness in the predictions of the GPT-4 model. We modified the system prompt (i.e., the annotation guidelines) with the aim of mitigating the issue. In order to answer RQ4, we analyzed the effects of the changes on the performance of the model. Specifically, we repeated all the preceding experiments with the modified prompt, and observed the changes in performance.

\subsection{Robustness (RQ5)}
The final experiment was focused on answering RQ5, that is, analyzing the robustness of the GPT-4 annotations. The preceding experiments yielded multiple sets of labels over the same data points. Each version of the annotation guidelines, that is, the original system prompt and the updated one, was associated with four labels for each data point---two from the single sentences experiments (labels only and labels with explanations), and two from the batch predictions. While these experiments differed in the form of how the model was prompted (i.e., with one or multiple sentences, and with or without an explanation), the annotation instructions remained the same. Therefore, this experiment explored how the form of the prompting affects the results. Specifically, we were interested in assessing stability of predictions across the four labels produced within different experiments relying on the same annotation guidelines.

\section{Results and Discussion}
\label{sec:results}

\subsection{GPT-4 Text Analysis (RQ1)}
\begin{figure*}[t]
  \centering
  \includegraphics[width=.246\linewidth]{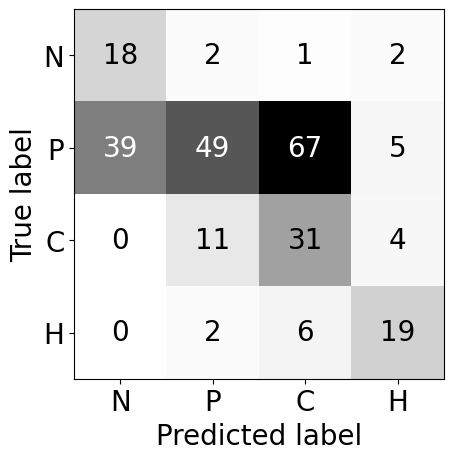}
  \includegraphics[width=.246\linewidth]{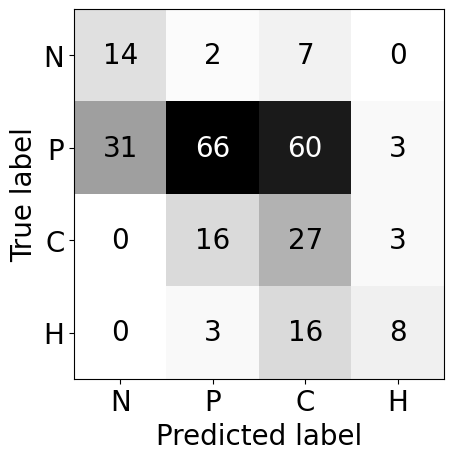}
  \includegraphics[width=.246\linewidth]{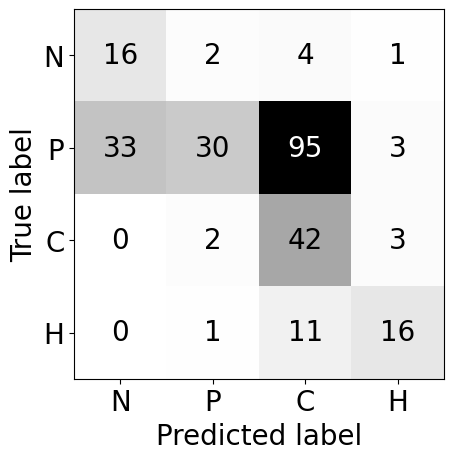}
  \includegraphics[width=.246\linewidth]{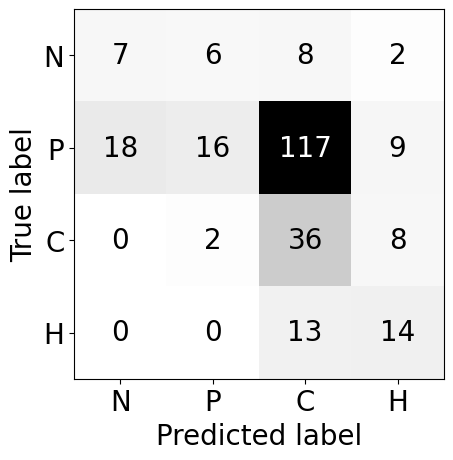}\\
  \includegraphics[width=.246\linewidth]{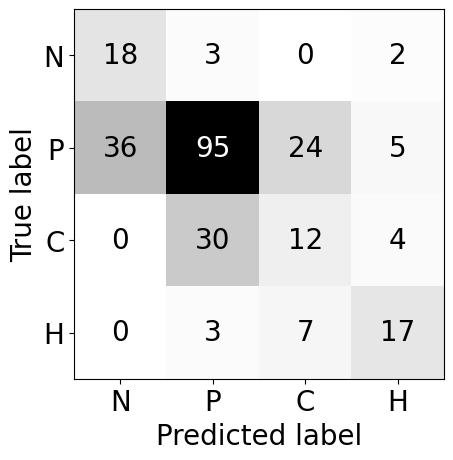}
  \includegraphics[width=.246\linewidth]{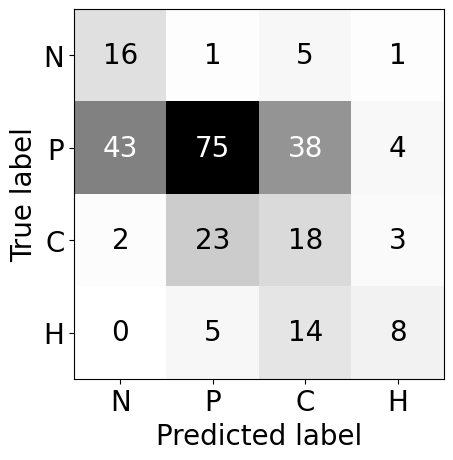}
  \includegraphics[width=.246\linewidth]{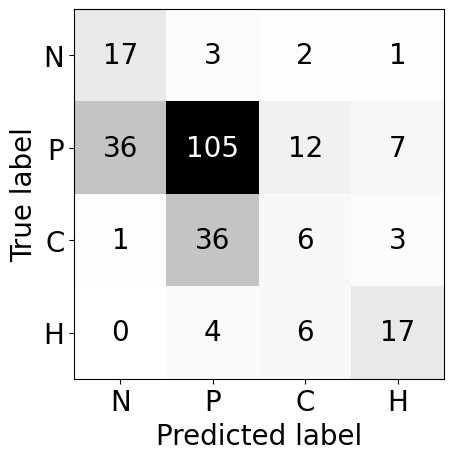}
  \includegraphics[width=.246\linewidth]{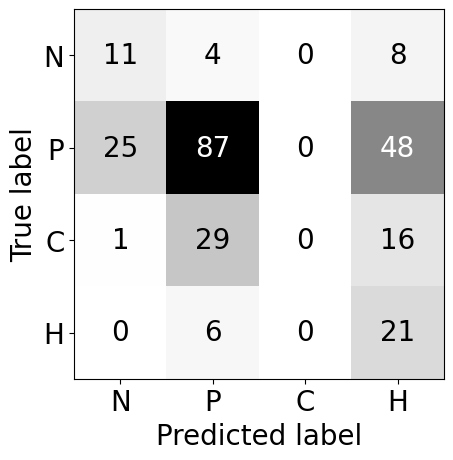}
  \caption{Confusion matrices for the original annotation guidelines (top row) and updated annotation guidelines (bottom row). From left to right the matrices describe the following experimental conditions: Single -- Labels Only, Batch -- Labels Only, Single -- Labels \& Explanation, Batch -- Labels \& Explanation. The labels: N -- No value, P -- Potential value, C -- Certain value, H -- High value.}
  \label{fig:cm}
\end{figure*}

The results of the experiment focused on GPT-4's performance on the text analysis task as compared to the human annotators (RQ1) are reported in Table \ref{tab:results} under the Original instructions and Single -- Labels Only entry. The overall F$_1=.53$ suggests that GPT-4 is able to successfully analyze the texts while at the same time leaving ample room for improvement. Additional insight is provided by the confusion matrix in the upper left corner of Figure \ref{fig:cm}. There, we can see that the system struggled with the Potential value label where many instances of this class were either predicted as No value or Certain value.

\begin{table*}[t]
  \caption{Experimental Results. The Instructions column encodes if the original or updated annotation guidelines were used in GPT-4's system prompt. The Annotation Modality column describes the experimental setting. The remaining columns report the performance metrics computed against the gold labels.}
  \label{tab:results}
  \begin{tabular}{llrrrrr}
    \toprule
    Instructions&Annotation Modality            &Precision&Recall&F1-score&Accuracy&$\alpha$\\
    \midrule
    Original    &Single -- Labels Only (RQ1)          &.63      &.46   &.53     &.46     &.51     \\
                &Batch -- Labels Only (RQ2)          &.61      &.45   &.52     &.45     &.42     \\
                &Single -- Labels \& Explanation (RQ3)&.69      &.40   &.51     &.40     &.44     \\
                &Batch -- Labels \& Explanation (RQ3) &.52      &.29   &.37     &.29     &.19     \\
    \midrule
    Updated     &Single -- Labels Only (RQ4)          &.60      &.55   &.57     &.55     &.53     \\
                &Batch -- Labels Only (RQ4)           &.57      &.46   &.51     &.46     &.42     \\
                &Single -- Labels \& Explanation (RQ4)&.58      &.57   &.57     &.57     &.48     \\
                &Batch -- Labels \& Explanation (RQ4) &.48      &.46   &.47     &.46     &.27     \\
  \bottomrule
\end{tabular}
\end{table*}

\begin{figure}[t]
  \centering
  \includegraphics[width=\linewidth]{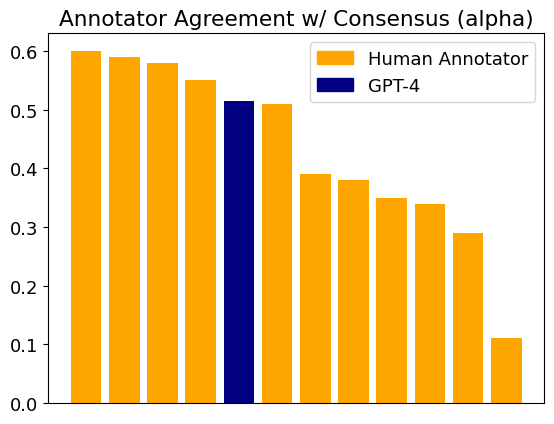}
  \caption{The annotator agreement (Krippendorff's $\alpha$) of the manually created annotations and GPT-4 predictions computed against the consensus (gold) labels. GPT-4 performs comparably to human annotators (law students).}
  \label{fig:alpha}
\end{figure}

It is important to recall that the task is very challenging even for human annotators and requires highly specialized domain expertise. Hence, we are interested in how the performance of GPT-4 compares to that of the human annotators. Figure \ref{fig:alpha} benchmarks the agreement, in terms of Krippendorff's $\alpha$, of GPT-4 with the consensus labels to the agreement of the law students' labels with the consensus. In Figure \ref{fig:alpha}, we can clearly recognize two groups of annotators, i.e., those whose agreements are $>.5$ and those whose agreements are $<.4$. This significant gap quite likely distinguishes between well-performing and less well-performing human annotators. GPT-4's performance is on par with the well-performing law student annotators.

\subsection{Batch Prediction (RQ2)}
The results of the experiment focused on GPT-4's performance on batch prediction (RQ2) are also reported in Table \ref{tab:results} under the Original instructions and Batch -- Labels Only entry. The overall F$_1=.52$ is a slight decrease in performance as compared to the prediction performed on one data point at a time. The significantly lower cost of this approach may justify the difference in performance. However, while the overall performance remained similar, the performance on the individual labels changed to a larger extent, as can be seen in the
corresponding confusion matrix shown in Figure \ref{fig:cm} (first row, second from the left). While the performance on the sentences with the Potential label is improved, the model performed less well on the sentences from the other three classes.

\subsection{Explanations -- CoT (RQ3)}
The results of the experiment focused on GPT-4's performance when providing explanations in addition to the predictions (RQ3) are reported in Table \ref{tab:results} under the Original instructions and Single -- Labels \& Explanation entry. Interestingly, we observe a decrease in performance as compared to the single sentence prediction experiment. The overall F$_1$ went from $0.53$ to $0.51$ and accuracy from $0.46$ to $0.40$. Further insight is provided by the confusion matrix in Figure \ref{fig:cm} (first row, second from the right). Apparently, the issue of predicting Potential value sentences as Certain value is even more pronounced than before. This strongly suggests that GPT-4 struggles with correctly interpreting  the annotation guidelines when it comes to distinguishing between the two classes. Note that this is contrary to the expectations of improving the performance by having GPT-4 explain its predictions since this is akin to CoT prompting which often leads to improvements in performance on a task.

The provided explanations often appear to be in agreement with the predicted labels but this is not always the case. For example, the following explanation is provided for a sentence that is correctly predicted as having No value:

\begin{quote}
    The sentence is a verbatim citation of the source provision and does not provide any additional information about the meaning of the phrase ``digital musical recording.''
\end{quote}

\noindent The following explanation is attached to a sentence that is wrongly predicted as Certain value:

\begin{quote}
    The sentence provides an explanation of what would not qualify under the basic definition of a digital musical recording, which is useful for understanding the boundaries of the phrase of interest.
\end{quote}

\noindent The sentence should have been predicted as High value and the explanation is aligned with such a prediction. Interestingly, it did not help to steer the prediction towards assigning the High value label.

The results of the batch experiment focused on further explorations of RQ3 are reported in Table \ref{tab:results} under the Original instructions and Batch -- Labels \& Explanation entry. We observe a complete degradation of the performance under this condition. As apparent from the corresponding confusion matrix in the upper right corner of Figure \ref{fig:cm}, a large portion of the sentences were mislabeled as having Certain value. This  suggests that the definition of the Certain value class may be too broad. Interestingly, the schematic explanations are generally in agreement with the predicted labels irrespective of the prediction being correct or not. Below are example predictions with explanations from the batch experiment:

\begin{verbatim}
Q1 Yes -> Q2 No -> Q4 Yes => High value
Q1 Yes -> Q2 No -> Q4 No -> Q5 Yes => Certain 
    value
Q1 Yes -> Q2 No -> Q4 No -> Q5 No => Potential 
    value
Q1 No => No value
\end{verbatim}

\noindent Recall that the Q\# refer to the questions from the annotation guidelines an annotator is supposed to consider in order to correctly label a sentence.

\subsection{Prompt (Annotation Guidelines) Modification (RQ4)}
The preceding experiments identified a potential issue with the definition of the Certain value class: it may be too broad. Hence, we use this particular issue as the test bed for investigating RQ4. Specifically, we modify the guidelines with the aim to mitigate the issue, i.e., improve the performance of the GPT-4 model on the task. The annotation guidelines contain the following definition of the Certain value class:

\begin{quote}
    The system should select this label if the sentence does not explicitly elaborate on the meaning of the phrase of interest, yet the sentence still provides grounds to draw some (even modest or quite vague) conclusions about the meaning of the phrase of interest.
\end{quote}

\noindent Furthermore, the guidelines direct an annotator to consider the below question after ruling out the High value and No value labels:

\begin{quote}
    Does the sentence provide useful context with respect to the elaboration of the meaning of the phrase of interest?
\end{quote}

\noindent A positive answer to that question should result in annotating the respective sentence with the Certain value label. A negative answer directs the annotator to assign the Positive value label. Indeed, the experiments focused on explanations clearly show that the system often tends to answer the question in positive. Consider the following example of an explanation in natural language:

\begin{quote}
    The sentence [...] does not explicitly elaborate on the meaning of the phrase ``cybercrime'' [...] However, it provides useful context by mentioning a convention that deals with cybercrime [...]
\end{quote}

\noindent Similarly, the following chain of reasoning is predominantly used in the batch prediction with explanation experiment (see Figure \ref{fig:user_msg_batch_exp} to understand the format of the below):

\begin{quote}
    Q1 Yes -> Q2 No -> Q4 No -> Q5 Yes
\end{quote}

\noindent Question 5 (Q5) is the one that directs an annotator to assign the sentence the Certain value label in case it is answered in positive.

Based on the above analysis, our aim is to modify the annotation guidelines to make the system less likely to annotate a sentence as Certain value and opt for a different label. To achieve this goal, we replaced the above definition of the Certain value class with a more restrictive one:

\begin{quote}
    The system should select this label if the sentence elaborates on the meaning of the phrase of interest implicitly.
\end{quote}

\noindent The definition follows up on the definition of the High value class where an \emph{explicit} elaboration is required. 





The results of the experiment focused on the effects of modifying the prompt (RQ4) are reported in Table \ref{tab:results} under the Updated instructions section. The overall F$_1=.57$ for the Single -- Labels Only condition is a noticeable improvement over the F$_1=.53$ performance with the original guidelines. The corresponding confusion matrix shown in the bottom left of Figure \ref{fig:cm} reveals that the issue of over-predicting the Certain class at the expense of the Potential value class has been addressed effectively. On the other hand, it appears that the system now errs on the other side, being reluctant to label a sentence as having Certain value. Nevertheless, the overall performance of the system appears to be improved.

Furthermore, application of the CoT prompting, i.e., asking the model to provide explanations alongside the predictions, no longer leads to dramatic deterioration of performance with the updated annotation guidelines. While we can still observe a slight decrease in performance of the CoT prompt for the batch prediction, it is quite small compared to the decrease observed with the original annotation guidelines.

\subsection{Robustness (RQ5)}
The results of the experiment focused on the robustness of GPT-4's predictions (RQ5) are reported in Table \ref{tab:robustness}. The table shows inter-annotator agreement (Krippendorff's $\alpha$) among the predictions from the earlier experiments. Interestingly, the agreement appears to be relatively low considering the fact that we are comparing systems based on the identical annotation guidelines. While further investigation is needed,  it appears that small changes in the expected format of the output can  dramatically affect the predictions.

\begin{table}[t]
  \caption{The inter-annotator agreement (Krippendorff's $\alpha$) between the predictions from the experiments (RQ5): S--LO: Single -- Labels Only, S--LE: Single -- Labels \& Explanation, B--LO: Batch -- Labels Only, B--LE: Batch -- Labels \& Explanation}
  \label{tab:robustness}
  \setlength{\tabcolsep}{1.5pt}
  \begin{tabular}{lrrrr@{\hspace{.4cm}}rrrr}
    \toprule
          &\multicolumn{4}{c}{Original}&\multicolumn{4}{c}{Updated}\\
          &S--LO&S--LE&B--LO&B--LE     &S--LO&S--LE&B--LO&B--LE\\
    \midrule
    S--LO &1.0  &.78  &.58  &.36       &1.0  &.83  &.55  &.37\\
    S--LE &     &1.0  &.48  &.44       &     &1.0  &.50  &.27\\
    B--LO &     &     &1.0  &.44       &     &     &1.0  &.58\\
    B--LE &     &     &     &1.0       &     &     &     &1.0\\
  \bottomrule
\end{tabular}
\end{table}

\section{Limitations}
In this study, we focused on a single specific task requiring highly specialized domain expertise, which may limit the generalizability of our findings. The task was selected based on the assumption that it represents the complex nature of tasks that may arise in specialized domains. However, it is possible that the performance of GPT-4 in other tasks requiring domain expertise might differ significantly. Moreover, the relatively small data set used in our analysis might not capture the full range of complexities and nuances associated with tasks requiring specialized knowledge. Consequently, the results obtained in this study should be interpreted with caution and not generalized to all tasks requiring domain expertise.

Another limitation concerns the general issues of reproducible experiments with proprietary OpenAI's GPT models. As access to these models is limited and often subject to certain terms and conditions, it can be challenging for independent researchers to replicate the experiments and validate the findings. This raises concerns about the reproducibility and robustness of the results, which are essential aspects of scientific research. Furthermore, any changes or updates to the GPT models by OpenAI might affect the performance and outcomes of experiments, making it difficult to establish a consistent baseline for comparison across studies. Therefore, it is crucial to address these concerns and develop strategies to promote reproducibility and robustness in future studies involving GPT models.

\section{Conclusions and Future Work}
This study assessed the capabilities of GPT-4 in analyzing textual data in the context of a task focused on interpretation of legal concepts. Our findings indicate that GPT-4 can perform at a level comparable to well-trained law student annotators. 
The fact that the model is able to take a multi-page document, understand the instructions contained therein, and apply these instructions to complex real-world textual data demonstrates the impressive performance of GPT-4. Further, this could have a significant impact on research in domains where complex annotation tasks are performed, such as the legal domain. Being able to utilize GPT-4, instead of hiring and training human annotators over extended periods of time could enable many types of research efforts, and open the door to novel large-scale research or data science projects.

We demonstrated that GPT-4 can be effectively utilized for batch predictions, offering significant cost reductions without a major decline in performance. On the other hand, CoT prompting did not yield a noticeable improvement in performance. We showcased an example of analyzing GPT-4's predictions to identify and address deficiencies in annotation guidelines, leading to improvements in the model's performance. However, the study also highlighted the model's brittleness, as minor formatting changes in the prompt had a substantial impact on the predictions. Researchers and practitioners can leverage these findings to effectively employ GPT-4 in semantic and pragmatic annotation tasks within specialized domains, while being mindful of the limitations.

Future work should focus on evaluation of GPT-4's capabilities across a broader range of tasks and domains, involving larger data sets, that require highly specialized expertise. Additionally, exploring methods to improve the model's robustness and resilience to minor formatting changes in the prompts would be valuable, ensuring more consistent and reliable performance. Furthermore, investigating alternative prompting techniques or fine-tuning strategies could potentially lead to enhanced performance in specialized tasks.
 \begin{acknowledgments}
 This work was supported in part by a National Institute of Justice Graduate Student Fellowship (Fellow: Jaromir Savelka) Award \# 2016-R2-CX-0010, “Recommendation System for Statutory Interpretation in Cybercrime,” a University of Pittsburgh Pitt Cyber Accelerator Grant entitled “Annotating Machine Learning Data for Interpreting Cyber-Crime Statutes,” and the National Science Foundation, grant no. 2040490, FAI: Using AI to Increase Fairness by Improving Access to Justice.
\end{acknowledgments}

\balance
\bibliography{sample-ceur}




\end{document}